\documentclass[10pt,twocolumn,letterpaper]{article}

\usepackage[pagenumbers]{cvpr}            %

\usepackage{multirow}
\usepackage{graphicx}
\usepackage[utf8]{inputenc}
\usepackage{algorithm}
\usepackage{makecell}
\usepackage{algpseudocode}
\usepackage{amsmath}
\usepackage{afterpage}
\usepackage{booktabs}
\usepackage{array}
\usepackage{graphicx}
\usepackage{colortbl}
\usepackage{alltt}
\usepackage{tcolorbox}
\usepackage{listings}
\usepackage[utf8]{inputenc}
\usepackage{fancyvrb}
\usepackage{fancyhdr}
\usepackage{caption}
\usepackage{multicol}

\usepackage{tocloft}

\usepackage{soul}

\usepackage{titletoc}
\usepackage{minitoc}
\definecolor{cvprblue}{rgb}{0.21,0.49,0.74}
\usepackage[pagebackref,breaklinks,colorlinks,allcolors=cvprblue]{hyperref}

\title{RAVEL: \ul{Ra}re Concept Generation and Editing \ul{v}ia Graph-driven R\ul{el}ational Guidance}

\author{
Kavana Venkatesh$^{1}$ \quad \quad
Yusuf Dalva$^{1}$ \quad \quad
Ismini Lourentzou$^{2}$ \quad \quad
Pinar Yanardag$^{1}$ \\
$^{1}$Virginia Tech \quad 
$^{2}$UIUC \\
\texttt{\{kavanav,ydalva, pinary\}@vt.edu} \quad \texttt{\{lourent2\}@illinois.edu} \\
\url{https://ravel-diffusion.github.io}
}

\begin{document}
\twocolumn[{
\maketitle
\begin{center}
    \captionsetup{type=figure}
    \vspace{-1em}
\newcommand{\imwidth}{1\textwidth}

\begin{tabular}{@{}c@{}}
 
\parbox{\imwidth}{\includegraphics[width=\imwidth, ]{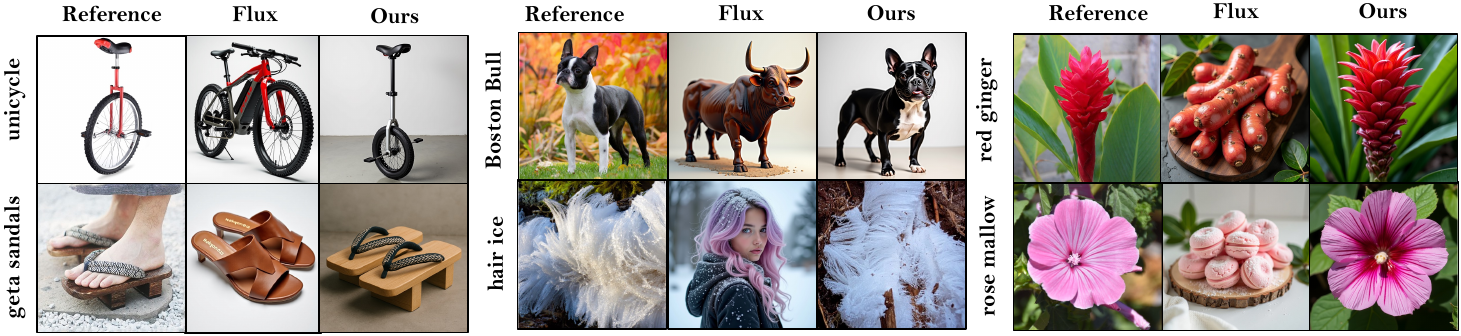}}
\\

\vspace{1em}
\end{tabular}
    \vspace{-2.5em}
    \captionof{figure}{We introduce \texttt{RAVEL}, a training-free approach that uses graph-based RAG to enhance T2I models with context-aware guidance. It improves generation of rare, complex concepts and supports disentangled image editing. A self-correction module further refines visual and narrative accuracy.} 
    \label{fig:teaser}
\end{center}
}]

\maketitle

\begin{abstract}
Despite impressive visual fidelity, current text-to-image (T2I) diffusion models struggle to depict rare, complex, or culturally nuanced concepts due to training data limitations. We introduce \texttt{RAVEL}, a training-free framework that significantly improves rare concept generation, context-driven image editing, and self-correction by integrating graph-based retrieval-augmented generation (RAG) into diffusion pipelines. Unlike prior RAG and LLM-enhanced methods reliant on visual exemplars, static captions or pretrained knowledge of models, RAVEL leverages structured knowledge graphs to retrieve compositional, symbolic, and relational context, enabling nuanced grounding even in the absence of visual priors. To further refine generation quality, we propose \textit{SRD}, a novel self-correction module that iteratively updates prompts via multi-aspect alignment feedback, enhancing attribute accuracy, narrative coherence, and semantic fidelity. Our framework is model-agnostic and compatible with leading diffusion models including Stable Diffusion XL, Flux, and DALL-E 3. We conduct extensive evaluations across three newly proposed benchmarks - \textit{MythoBench}, \textit{Rare-Concept-1K}, and \textit{NovelBench}. \texttt{RAVEL} also consistently outperforms SOTA methods across perceptual, alignment, and LLM-as-a-Judge metrics. These results position \texttt{RAVEL} as a robust paradigm for controllable and interpretable T2I generation in long-tail domains.
\end{abstract}
    
\section{Introduction}
\label{sec:intro}
Recent advances in T2I diffusion models \cite{ho2020denoising, song2020denoising, saharia2022photorealistic, nichol2021glide, podell2023sdxl} have led to striking progress in generating visually rich imagery from natural language prompts. However, these models remain inherently limited by the scope and distribution of their training data, which tends to underrepresent rare, complex, or culturally nuanced concepts \cite{yamaguchi2023limitation}. As a result, T2I models often fail to accurately depict entities or scenarios that fall outside common visual domains, ranging from real-world long-tail concepts like a `unicycle' or `Boston bull' to intricate historical figures, symbolic relationships, and imaginative fictional constructs \cite{samuel2024generating}. This highlights a critical bottleneck in the generalizability and contextual grounding of current generative models when faced with sparsely represented or visually undocumented concepts.

To overcome these limitations, users often resort to writing highly detailed prompts to guide the models toward the desired outputs \cite{witteveen2022investigating, wang2022diffusiondb, liu2022design}. However, this process is time-consuming and requires extensive manual trial-and-error, domain expertise, as well as understanding of the model’s specific vocabulary and biases. Another approach is to fine-tune models on specialized datasets, typically by training new low-rank adaptations (LoRAs) \cite{hu2021lora} or creating custom embeddings \cite{gal2022image} for each unique concept. Although effective, these methods require significant computational resources, time, and technical expertise, making them costly and impractical for users without advanced skills. Recent approaches like prompt engineering, adapter modules, and few-shot learning have aimed to extend model capabilities without full-scale retraining \cite{zhang2022promptgen, Yao2024promptcot, aws_rag_blog}, yet they depend heavily on extensive user input and still work within the boundaries of the model's existing knowledge. Fine-tuning LLMs on domain-specific data for contextual guidance is resource intensive, prone to issues like hallucination and catastrophic forgetting \cite{french1999catastrophic}, and often struggles to capture fine-grained and relational details accurately \cite{ghosh2024closer}. 

Retrieval-Augmented Generation (RAG)  \cite{lewis2020retrieval} techniques have been introduced to expand context in large language models (LLMs) \cite{touvron2023llama}, leveraging external knowledge to improve generation consistency and depth. Prior RAG-based methods \cite{chen2022re, blattmann2022ragdiffusion, shrestha2024fairrag, guo2024refir, tang2024retribooru, aws_rag_blog, lu2024recon, shalev2025imagerag, lyu2025realrag} augment generation using retrieved image-text pairs, captioned concepts, or diffusion trajectories from large-scale databases. While effective for reference-based style transfer, these approaches (1) assume the existence of visually similar examples, (2) lack fine-grained or relational reasoning, and (3) operate in a single-pass, non-iterative fashion. In contrast, \textbf{our method is the first to use structured knowledge graphs as the retrieval backbone}, enabling symbolic, relational, and factual grounding without requiring visual exemplars. Moreover, our novel \texttt{SRD} module introduces an iterative self-correction loop that dynamically identifies missing attributes and issues targeted refinements, which is absent in all prior RAG-T2I methods. 


To address these limitations, we propose a novel framework that augments T2I diffusion models with a graph-based RAG system. Our approach dynamically retrieves structured textual context from a knowledge graph, enabling diffusion models to generate semantically rich, contextually grounded images, even for rare or unseen concepts beyond their training data. This integration enhances not only the accuracy and consistency of generations, but also their interpretability and cultural fidelity without requiring additional user input or fine-tuning. The framework is model-agnostic and compatible with popular diffusion systems, including Flux \cite{esser2024scaling}, Stable Diffusion \cite{rombach2022high}, DALL·E \cite{dalle3_2023}, and ControlNet \cite{zhang2023adding}. Beyond generation, it also improves image editing within ControlNet by injecting contextual cues into the editing prompt, enabling finer control and semantic precision. Additionally, we introduce a self-correction mechanism (SRD) that iteratively refines outputs based on vision-language alignment across key dimensions such as appearance, narrative coherence, and cultural relevance. Unlike prior self-correcting methods \cite{wu2024self, zhuo2025reflection} that rely on the static internal knowledge of pretrained LLMs, our RAG-driven loop dynamically expands the contextual window, allowing for structured, multi-aspect refinement. This leads to significantly improved fidelity and semantic alignment, pushing the boundaries of interpretable and controllable generative modeling. Our contributions are as follows:

\begin{itemize}
    \item We propose \texttt{RAVEL}, a framework that augments text-to-image diffusion models by dynamically retrieving relevant information from a structured knowledge graph to generate complex rare concepts across diverse domains. To our knowledge, \texttt{RAVEL} represents the first instance of leveraging Knowledge Graph-Based RAG to improve both image generation and image editing. 
    \item We propose a novel RAG context-driven self-correction approach for image generation, refining outputs iteratively by using structured knowledge to address inaccuracies, achieving coherence in complex visual narratives.
    \item We release three new benchmarks - \textit{MythoBench, Rare-Concept-1K}, and \textit{NovelBench}, designed to rigorously evaluate generative models on cultural symbolism, fine-grained attribute grounding, and long-tail generalization. These resources provide the first comprehensive testbeds for rare concept generation and editing under structured, multi-aspect evaluation.
\end{itemize}

\section{Related Work}
\label{sec:related}

\noindent \textbf{Large Language Models and RAG} Large Language Models (LLM) have shown remarkable human-like general text generation abilities for a variety of tasks \cite{brown2020language, dalle3_2023}, further enhanced by prompt engineering \cite{wei2022chain, brown2020language, chia2023contrastive, yao2023tree, wang2022self}. Retrieval Augmented Generation (RAG) \cite{lewis2020retrieval} is a prompt engineering technique that improves the relevance and accuracy of a generative model's response, while reducing hallucinations \cite{bechard2024reducing}. Knowledge Graphs (KG) consist of consistent nodes and relationships to store information in a structured way. KGs are extensively being used as databases for RAG frameworks \cite{sen2023knowledge, peng2024graph} due to their scalability, interpretability and extendability. In our work, we leverage KGs to store structured and rich contextual data to enhance T2I diffusion model prompts.

\noindent \textbf{LLMs and RAG for T2I Diffusion Models} Advancements in large language models (LLMs) have enhanced text-to-image (T2I) diffusion models, improving image fidelity and prompt alignment. Recent approaches have leveraged LLMs to generate layouts with bounding boxes for better spatial organization in complex, multi-object scenes \cite{gani2023llm, feng2023layoutgpt, lian2023llm, Feng_2024_CVPR}. Prompt engineering also extends T2I models’ narrative understanding \cite{zhang2022promptgen}, while fine-tuning LLMs on high-quality prompts improves output \cite{Yao2024promptcot} but can be costly and less adaptable, with risks of forgetting and hallucination.
Retrieval-augmented generation (RAG) \cite{lewis2020retrieval} further enhances T2I models by retrieving relevant images or prompts from vector databases, aiding in style and concept alignment \cite{chen2022re, blattmann2022ragdiffusion, shrestha2024fairrag, guo2024refir, tang2024retribooru, aws_rag_blog, shalev2025imagerag, lu2024recon, lyu2025realrag}. However, these approaches remain constrained by their reliance on retrieved visual exemplars and single-pass pipelines. They lack the ability to reason over fine-grained, structured knowledge, making them less effective for rare concept generation, context fidelity, and self-correcting edits, capabilities central to our proposed method.
LLM-controlled methods in image editing, like DiffEdit \cite{couairon2022diffedit} for targeted edits and Hive \cite{Zhang_2024_CVPR} for compositional edits, show promise but struggle with domain consistency. Techniques such as InstructPix2Pix \cite{brooks2023instructpix2pix} and Prompt2Prompt \cite{hertz2022prompt} facilitate global edits but lack precision in finer details, highlighting the limitations in meeting nuanced, context-sensitive editing needs.

\section{Method}
\label{sec:method}

\subsection{Knowledge Graph Construction}

We employ a knowledge graph-based RAG framework to enhance T2I models to generate rare concepts beyond their pretraining distribution, without relying on visual priors. Unlike unstructured retrieval, knowledge graphs encode entities, attributes, and relationships explicitly, enabling precise, compositional, and interpretable context. Our domain-agnostic graph construction follows a two-phase automated pipeline. Given a domain template, we scrape high-quality data (e.g., Wikipedia, verified sources) and apply Chain-of-Thought \cite{wei2022chain} reasoning and Few-Shot prompting \cite{brown2020language} to extract structured, attribute-rich descriptions - e.g., taxonomy and habitat for species, or appearance and symbolic artifacts for cultural figures. In phase one, we create per-entity graphs using LLM-generated graph queries; in phase two, we form bidirectional links to support relational reasoning (e.g., associating hummingbirds via pollination traits or referencing historical events). This modular pipeline generalizes across domains by adapting data sources and attribute schemas. 

\subsubsection{Graph-Based RAG}
\label{subsec:Graph-based-RAG}

Consider a simple user prompt $\mathcal{P}$ such as ``red ginger blooming in a garden". Our approach first leverages an LLM to identify core entities and related concepts; e.g., botanical species, scene setting, and potential relational cues. Once key entities are extracted, we employ an LLM to dynamically generate graph queries to retrieve their associated attributes and relationships from the knowledge graph. This yields a structured context set that encompasses both appearance features (e.g., bract shape, color, arrangement) and relational cues (e.g., pollinator = hummingbird, habitat = tropical forests). The retrieved context is then expanded using Contrastive CoT prompting \cite{chia2023contrastive} to highlight multiple visual and narrative dimensions, creating a richly grounded prompt that would be tedious to manually author. 

\noindent \textbf{Image Generation:} Given a high-level user prompt \(\mathcal{P}\), we generate a context-enriched prompt \(\mathcal{P'}\) by leveraging knowledge graph-based RAG and feed it to the T2I model. By enriching the initial prompts with extensive contextual and appearance information, \texttt{RAVEL} enables the generation of accurate visual representations. 

\noindent \textbf{Image Editing:}  Given an input image and an editing prompt   $\mathcal{P}_e$ such as ``Add the pollinator of red ginger", \texttt{RAVEL} queries the knowledge graph to retrieve precise contextual details; e.g., pollinators like hummingbird, orchid bee, or butterfly. This ensures the accurate and semantically grounded addition of the correct entity to the image, in contrast to standard ControlNet, which may insert a generic or unrelated object due to the lack of symbolic guidance.

\subsection{Self-Correcting RAG-Guided Diffusion (SRD)}

While our graph-augmented generation significantly improves accuracy, complex rare concepts with fine-grained attributes may still contain errors or omissions. To address this, we introduce \textbf{S}elf-Correcting \textbf{R}AG-Guided \textbf{D}iffusion (\textbf{SRD}), an iterative refinement mechanism that progressively corrects these issues by analyzing generated images and updating prompts based on graph context.

\noindent \textbf{Initialization and Setup.} Given the initial zero-shot image \( I_0 \), generated from the enhanced prompt \( P_0 \) (Section~\ref{subsec:Graph-based-RAG}), SRD compares \( I_0 \) to target features \( T \), derived from the RAG context to identify misalignments. We define two alignment metrics: a per-feature \textit{Stability Score} $s_i \in [0,1]$ indicating correctness of feature \( i \), and the \textit{Global Stability Index}, \(GSI\), which averages scores across all features. SRD is triggered when \( GSI(I_0) < GSI_{\epsilon} \). To refine the image, SRD iteratively adjusts prompts using RAG context guidance. A decay factor \( d \), (starting at 0.9, decreasing each round) scales prompt adjustments at each iteration \( k \), preventing overcorrection while allowing aggressive early fixes. A \textit{Change Tracking Matrix}, $S_{\text{tracker}}$ records stability counts for each feature across \(K \) iterations. Features stable for \(N \) consecutive rounds are `locked' to prevent regression. The process terminates when $GSI \geq GSI_{\epsilon}$,or when iteration limit \(K \) is reached.

\noindent \textbf{Iterative Refinement and Stability Tracking.} The SRD loop executes for up to \( K \) iterations, refining prompts and regenerating images to improve fidelity. At each step \( k \), image \( I_k \) is generated from prompt \( P_{k-1} \). An LLM analyzes \( I_k \) to assess current features \( F_{\text{curr}} \), computing stability scores in \( S_{\text{tracker}} \) based on alignment with target features \( T \). For instance, in the case of a red ginger plant, the system evaluates visual features such as bract structure, arrangement, and glossiness. If any attributes are missing, hallucinated, or regressing, the prompt is refined by adjusting contextual emphasis, modulated by a decay factor, \(d\). If \( F_{\text{curr}} = T \) and all feature scores meet the stability threshold \( S_{\text{threshold}} \), the process halts early due to sufficient alignment. Through this iterative refinement, SRD adapts its prompt refinement process to each feature’s stability status, creating a feedback loop that converges on high fidelity without excessive intervention. 


\begin{figure*}[t!]
    \centering
    \includegraphics[width=0.9\textwidth]{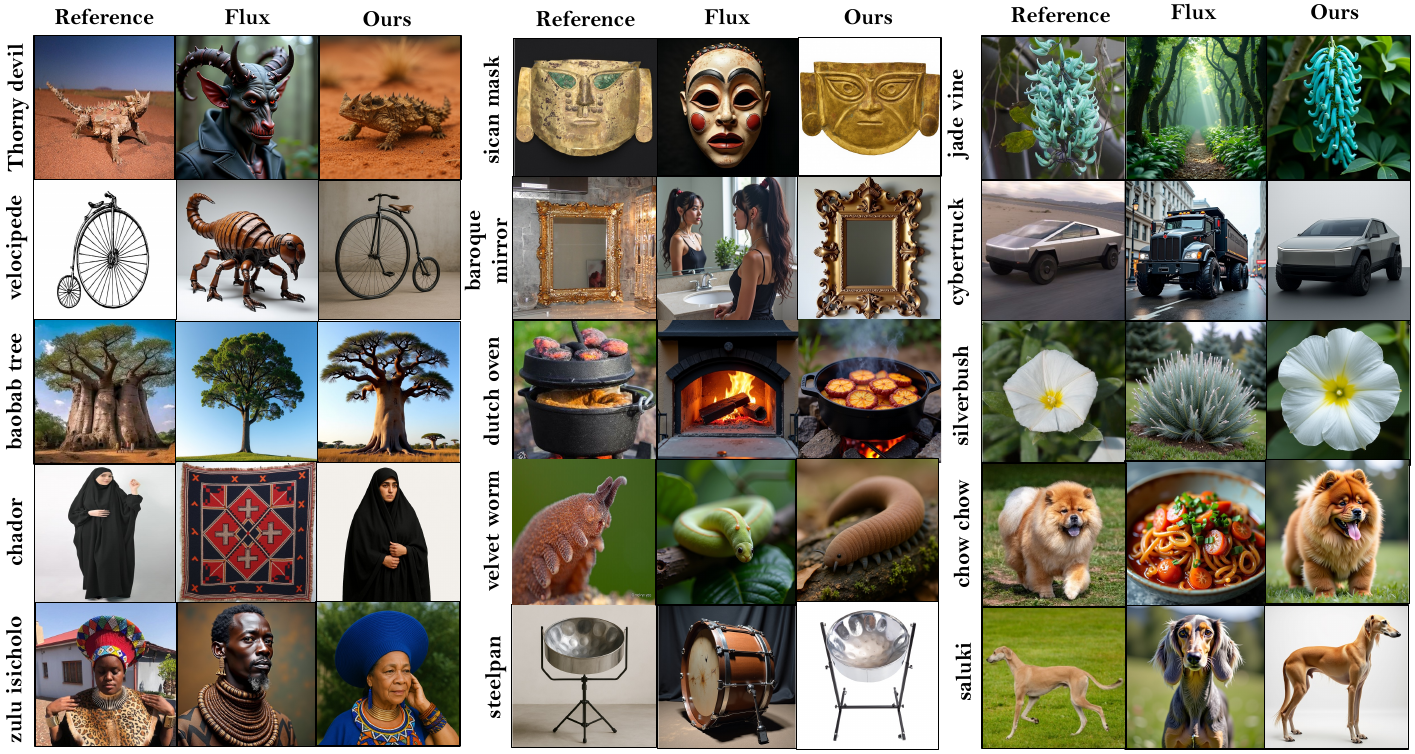} 
    \caption{RAVEL enhances image generation by integrating contextual details often overlooked by standard models for a variety of domains. \textit{\textbf{Note that the reference images  are shown solely for illustrative purposes and are not used by our framework. }}}
    \label{fig:qualitative-flux}
\end{figure*}

 \section{Experiments}
\label{sec:experiments}

We evaluate \texttt{RAVEL} across a broad range of qualitative and quantitative experiments to assess its performance in rare concept generation, contextual editing, and iterative self-correction. Our framework is tested across three state-of-the-art text-to-image (T2I) models - such as Flux, Stable Diffusion XL (SDXL), and DALL-E 3.

\noindent \textbf{Datasets and Domains:} We conduct experiments across more than 15 diverse rare concept categories, including rare animals, birds, fashion elements, architectural artifacts, symbolic cultural figures, and complex mythological characters. Our dataset is curated from a wide range of validated sources, including Wikipedia, original mythological texts, and narrative-rich domains from Project Gutenberg. Each concept is enriched using structured, multi-attribute prompting from a knowledge graph, enabling context-aware generation across domains. 

\noindent \textbf{Benchmarks:} We report performance on 100 relatively rare image prompts carefully chosen from the T2I-CompBench \cite{NEURIPS2023_f8ad010c}, an established benchmark that evaluates compositional and attribute-based accuracy in T2I generation. 
To address gaps in existing evaluation protocols and enable principled assessment of rare concept generation, we introduce \textbf{three novel benchmarks}, each designed to target a specific axis of rarity, generalization, and symbolic complexity:

\begin{itemize}
    \item \textbf{MythoBench:} The first benchmark for mythological and culturally symbolic concepts, containing 500 character prompts across three difficulty tiers (simple, composite, relational) based on narrative depth \& visual complexity.

    \item \textbf{Rare-Concept-1K:} A 1,000-prompt benchmark spanning five underrepresented categories - obsolete technologies, rare creatures, cultural artifacts, architectural objects, and uncommon tools. Each includes 200 manually verified concepts with structured contextual descriptions to evaluate attribute fidelity, domain-specific grounding.

    \item \textbf{NovelBench:} A zero-shot benchmark of 500 prompts sourced from \texttt{Project Gutenberg} (e.g., folklore, speculative fiction, period literature), designed to test generalization to unseen, narrative-rich, and archaic concepts under distribution shift.
\end{itemize}

\noindent
These benchmarks address the lack of reliable ground-truth imagery in rare domains and support comprehensive evaluation across symbolic reasoning, real-world rarity, and narrative generalization. We use four LLM-as-a-Judge \cite{zheng2023llm_as_a_judge} metrics: \texttt{Attribute Accuracy}, \texttt{Context Relevance}, \texttt{Visual Fidelity}, and \texttt{Intent Representation}. 

\noindent \textbf{Experimental Setup:} For data curation, we leverage Neo4j \cite{neo4j} as our knowledge graph database storing data of more than 600 distinct rare concepts and utilize GPT-4o as our LLM for all tasks. Each prompt is enriched with additional context retrieved by our RAG framework using Contrastive CoT prompting for output rare image generation.
To evaluate these generated outputs, we sample 500 prompts from the above four benchmarks and generate the 500 images each using three models-Stable Diffusion XL Base 1.0, Flux.1 [dev], and DALL-E 3, under both base prompt conditions (only name of a concept) and context-enhanced conditions across the 15+ categories. For image editing and self-correction, we assess our framework on 150 images featuring 38 distinct characters. To effectively capture fine-grained details, we set the guidance scale within a range of 15 to 30. $GSI_{\epsilon}$ is set to 23 for SRD. All experiments are conducted on a single NVIDIA L40 GPU.
For evaluation, we employ a multi-faceted approach to assess both our RAG system and the generated images. Given the complexity of domain-specific fine-grained details, we adopt the LLM-as-a-Judge framework \cite{zheng2023llm_as_a_judge}, implemented using G-Eval \cite{liu2023g} from the open-source DeepEval \cite{deepeval} library across our four benchmarks to isolate category-based performance. Additionally, we compute traditional computer vision metrics to quantify image quality and alignment across concepts sampled from the four benchmarks as mentioned and compare \texttt{RAVEL} with several SOTA methods for rare image generation, relational editing and self-correction. 

\subsection{Qualitative Results}  
\label{sec:qualitative-results}

\texttt{RAVEL} also effectively generates complex global mythology and fictional concepts without prior visual exemplars. 

\begin{figure*}[t!]
    \centering
    \includegraphics[width=0.85\textwidth]{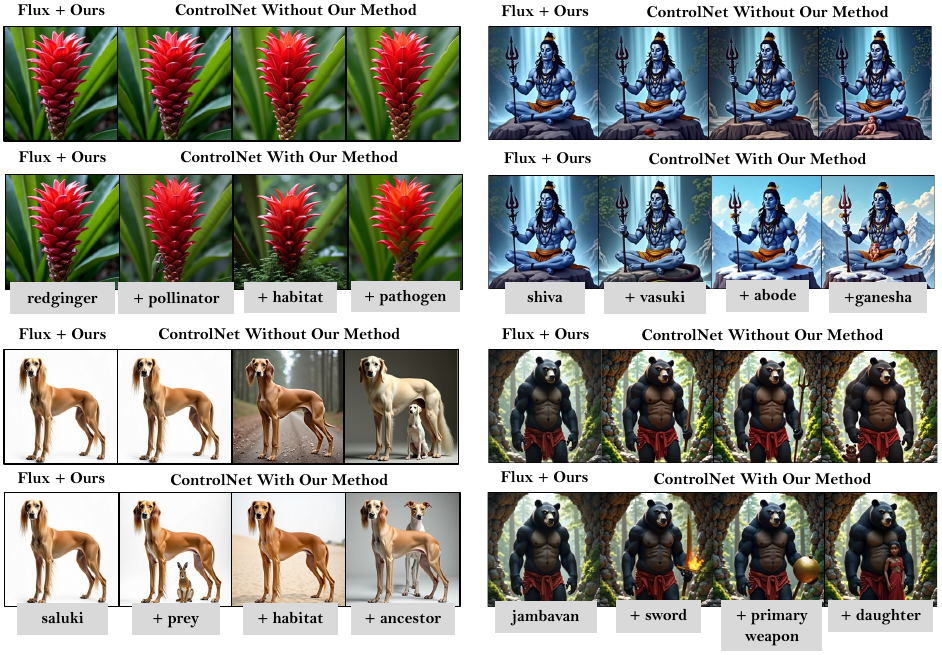} %
    \caption{Our method enhances  disentangled editing by adding relationally accurate elements without explicit instructions, while   ControlNet either adds generic objects or fails to make any edit. 
    }
    \label{fig:qual-editing}
\end{figure*}

\begin{table*}[t]
\centering
\caption{LLM-as-a-Judge Evaluation Across Multiple Benchmarks. \texttt{RAVEL} significantly improves alignment, particularly on knowledge-intensive datasets. *$\Delta_{avg}$ (change in average performance).}
\resizebox{\textwidth}{!}{%
\begin{tabular}{l|l|c|c|c|c|c|c|c}
\toprule
\textbf{Benchmark} & \textbf{Metric} & \textbf{SDXL} & \textbf{Flux} & \textbf{DALL-E3} & \textbf{SDXL+Ours} & \textbf{Flux+Ours} & \textbf{DALL-E+Ours} & $\Delta_{\text{avg}}$ \\
\midrule

\multirow{3}{*}{\textbf{T2I-CompBench}\cite{NEURIPS2023_f8ad010c}}
& Comp. Accuracy & 0.5120 $\pm$ 0.009 & 0.5870 $\pm$ 0.007 & 0.6230 $\pm$ 0.011 & 0.7210 $\pm$ 0.012 & 0.7980 $\pm$ 0.009 & 0.8320 $\pm$ 0.012 & \textbf{+0.206} \\
& CLIP-I         & 0.2730 $\pm$ 0.010 & 0.3180 $\pm$ 0.012 & 0.3510 $\pm$ 0.011 & 0.6620 $\pm$ 0.010 & 0.7010 $\pm$ 0.011 & 0.7280 $\pm$ 0.012 & \textbf{+0.189} \\
& BLIP-VQA       & 0.2120 $\pm$ 0.011 & 0.2510 $\pm$ 0.010 & 0.2790 $\pm$ 0.009 & 0.4240 $\pm$ 0.010 & 0.4530 $\pm$ 0.011 & 0.4680 $\pm$ 0.010 & \textbf{+0.215} \\
\midrule

\multirow{5}{*}{\textbf{MythoBench}}
& Attr. Accuracy   & 0.3470 $\pm$ 0.011 & 0.4070 $\pm$ 0.006 & 0.5000 $\pm$ 0.008 & 0.7790 $\pm$ 0.010 & 0.8650 $\pm$ 0.013 & 0.8820 $\pm$ 0.010 & \textbf{+0.456} \\
& Context Relevance& 0.4570 $\pm$ 0.009 & 0.5530 $\pm$ 0.007 & 0.5600 $\pm$ 0.014 & 0.8490 $\pm$ 0.011 & 0.8840 $\pm$ 0.011 & 0.9240 $\pm$ 0.014 & \textbf{+0.341} \\
& Intent Repres.   & 0.3320 $\pm$ 0.009 & 0.3160 $\pm$ 0.010 & 0.5120 $\pm$ 0.014 & 0.8090 $\pm$ 0.012 & 0.8990 $\pm$ 0.007 & 0.8540 $\pm$ 0.014 & \textbf{+0.496} \\
& Visual Fidelity  & 0.4190 $\pm$ 0.010 & 0.5040 $\pm$ 0.009 & 0.5390 $\pm$ 0.010 & 0.7620 $\pm$ 0.011 & 0.8120 $\pm$ 0.010 & 0.8290 $\pm$ 0.012 & \textbf{+0.304} \\
& Overall Score    & 0.3888 $\pm$ 0.023 & 0.4450 $\pm$ 0.023 & 0.5278 $\pm$ 0.018 & 0.7998 $\pm$ 0.015 & 0.8650 $\pm$ 0.017 & 0.8723 $\pm$ 0.018 & \textbf{+0.393} \\
\midrule

\multirow{5}{*}{\textbf{Rare-Concept-1K}}
& Attr. Accuracy   & 0.3980 $\pm$ 0.010 & 0.4670 $\pm$ 0.007 & 0.5230 $\pm$ 0.010 & 0.7410 $\pm$ 0.012 & 0.8120 $\pm$ 0.009 & 0.8450 $\pm$ 0.014 & \textbf{+0.348} \\
& Context Relevance& 0.4210 $\pm$ 0.014 & 0.5010 $\pm$ 0.007 & 0.5480 $\pm$ 0.012 & 0.6980 $\pm$ 0.010 & 0.7890 $\pm$ 0.011 & 0.8210 $\pm$ 0.014 & \textbf{+0.279} \\
& Intent Repres.   & 0.3890 $\pm$ 0.008 & 0.4450 $\pm$ 0.014 & 0.5370 $\pm$ 0.012 & 0.7120 $\pm$ 0.010 & 0.8010 $\pm$ 0.014 & 0.8340 $\pm$ 0.009 & \textbf{+0.312} \\
& Visual Fidelity  & 0.4350 $\pm$ 0.009 & 0.5010 $\pm$ 0.010 & 0.5340 $\pm$ 0.011 & 0.7390 $\pm$ 0.010 & 0.7940 $\pm$ 0.009 & 0.8180 $\pm$ 0.010 & \textbf{+0.290} \\
& Overall Score    & 0.4108 $\pm$ 0.019 & 0.4785 $\pm$ 0.018 & 0.5355 $\pm$ 0.017 & 0.7225 $\pm$ 0.013 & 0.7990 $\pm$ 0.015 & 0.8295 $\pm$ 0.014 & \textbf{+0.309} \\
\midrule

\multirow{5}{*}{\textbf{NovelBench}}
& Attr. Accuracy   & 0.4210 $\pm$ 0.014 & 0.5120 $\pm$ 0.010 & 0.5670 $\pm$ 0.006 & 0.7650 $\pm$ 0.007 & 0.8430 $\pm$ 0.010 & 0.8780 $\pm$ 0.010 & \textbf{+0.334} \\
& Context Relevance& 0.4480 $\pm$ 0.006 & 0.5340 $\pm$ 0.014 & 0.5890 $\pm$ 0.014 & 0.8120 $\pm$ 0.011 & 0.8710 $\pm$ 0.009 & 0.9010 $\pm$ 0.014 & \textbf{+0.319} \\
& Intent Repres.   & 0.4120 $\pm$ 0.013 & 0.4780 $\pm$ 0.011 & 0.5430 $\pm$ 0.007 & 0.7890 $\pm$ 0.007 & 0.8560 $\pm$ 0.012 & 0.8670 $\pm$ 0.006 & \textbf{+0.382} \\
& Visual Fidelity  & 0.4320 $\pm$ 0.010 & 0.5070 $\pm$ 0.009 & 0.5510 $\pm$ 0.010 & 0.7580 $\pm$ 0.010 & 0.8260 $\pm$ 0.011 & 0.8510 $\pm$ 0.009 & \textbf{+0.305} \\
& Overall Score    & 0.4283 $\pm$ 0.012 & 0.5078 $\pm$ 0.015 & 0.5625 $\pm$ 0.011 & 0.7810 $\pm$ 0.006 & 0.8490 $\pm$ 0.010 & 0.8740 $\pm$ 0.013 & \textbf{+0.327} \\
\bottomrule
\end{tabular}
}
\label{tab:llmjudge-benchmarks}
\end{table*}

\noindent \textbf{Image Editing:} Figure~\ref{fig:qual-editing} demonstrates how \texttt{RAVEL}, when integrated with ControlNet, enables precise and context-aware image editing using relational cues. We compare edits on three types of entities - plants (\textit{red ginger}), animals (\textit{saluki}), and mythological characters (\textit{Shiva, Jambavan}) under attribute-driven prompts (e.g., \textit{+pollinator}, \textit{+habitat}, \textit{+daughter}). Without our method, ControlNet often fails to apply meaningful edits or resorts to generic insertions. In contrast, \texttt{RAVEL} accurately grounds edits in concept-specific relational knowledge, such as pairing Shiva with his snake companion \textit{Vasuki} or placing red ginger in its tropical forest habitat, and adding hummingbird as its pollinator. Our method consistently adds narratively coherent, visually distinct, and semantically grounded elements without requiring explicit visual supervision. 

\begin{figure}[t!]
    \centering
    \includegraphics[width=\columnwidth]{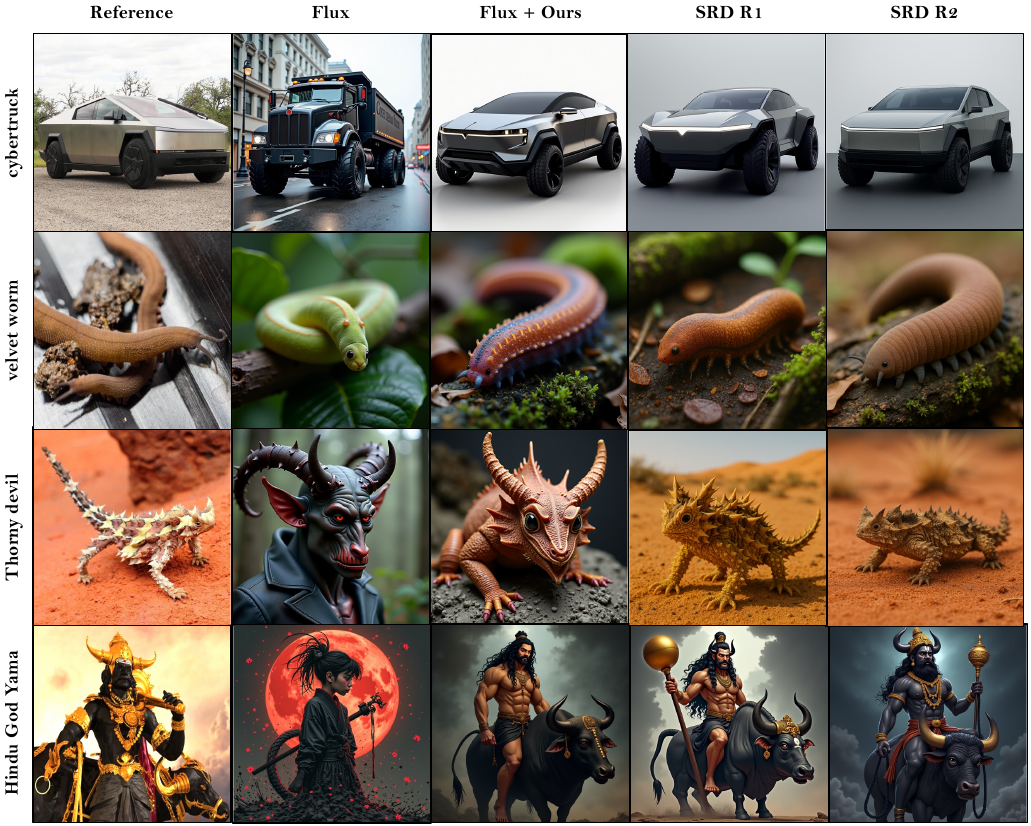} %
    \caption{Our self-correction mechanism ensures  accurate depictions of concepts via iterative, context-aware prompt refinement.}
    \label{fig:correcting}
\end{figure}

\noindent \textbf{Self-Correction for Image Alignment:} 
Figure~\ref{fig:correcting} illustrates how our SRD module incrementally improves image fidelity through iterative, context-aware prompt refinement. Starting from an initial generation (Flux + Ours), SRD analyzes visual deviations relative to the target attributes retrieved from the knowledge graph and refines the prompt over successive rounds (R1, R2). This allows SRD to correct both fine-grained structure and high-level semantics. For example, SRD sharpens the geometric body contours of the \textit{cybertruck}, fixes segmentation and texture inconsistencies in the \textit{velvet worm}, restores accurate spines and proportions in the \textit{thorny devil}, and adjusts appearance traits such as posture, accessories, and symbolic objects for \textit{Yama}. Compared to prior self-editing approaches \cite{wu2024self, zhuo2025reflection}, which depend solely on an LLM’s internal knowledge, SRD leverages retrieved, concept-specific context and thus produces more stable and interpretable corrections. The refinement process is model-agnostic and applies consistently across T2I architectures, including Flux, SDXL, and DALL-E.

\begin{table*}[h]
\caption{\textbf{Performance on Rare Concept Generation.} Our method significantly improves text-image alignment and attribute accuracy across multiple diffusion backbones.}
\centering
\tiny
\setlength{\tabcolsep}{4pt}
\renewcommand{\arraystretch}{1.2}
\resizebox{\textwidth}{!}{%
\begin{tabular}{l|c|c|c|c|c|c}
\hline
\textbf{Method} & 
\textbf{CLIP-T} $\uparrow$ & 
\textbf{CLIP-I} $\uparrow$ & 
\textbf{DINO v2} $\uparrow$ & 
\textbf{BLIP-VQA} $\uparrow$ & 
\textbf{LPIPS} $\downarrow$ & 
\textbf{SigLIP} $\uparrow$ \\
\hline

\multicolumn{7}{c}{\textbf{Baseline Models}} \\
\hline
Flux & 0.1871 $\pm$ 0.012 & 0.4178 $\pm$ 0.025 & 0.2871 $\pm$ 0.019 & 0.2778 $\pm$ 0.018 & 0.5802 $\pm$ 0.031 & 0.0634 $\pm$ 0.007 \\
SDXL & 0.1842 $\pm$ 0.013 & 0.2971 $\pm$ 0.021 & 0.2619 $\pm$ 0.017 & 0.1994 $\pm$ 0.016 & 0.6234 $\pm$ 0.034 & 0.0335 $\pm$ 0.005 \\
DALL-E & 0.2201 $\pm$ 0.014 & 0.4802 $\pm$ 0.027 & 0.3667 $\pm$ 0.022 & 0.2789 $\pm$ 0.020 & 0.5677 $\pm$ 0.030 & 0.0689 $\pm$ 0.008 \\
\hline

\multicolumn{7}{c}{\textbf{RAG-Based Methods}} \\
\hline
ImageRAG \cite{shalev2025imagerag} & 0.2841 $\pm$ 0.026 & 0.5175 $\pm$ 0.011 & 0.2877 $\pm$ 0.002 & 0.2162 $\pm$ 0.0012 & 0.6538 $\pm$ 0.030 & 0.0420 $\pm$ 0.010 \\
*RDM \cite{blattmann2022ragdiffusion} & 0.2985 $\pm$ 0.021 & 0.5102 $\pm$ 0.010 & 0.3017 $\pm$ 0.010 & 0.2345 $\pm$ 0.013 & 0.5908 $\pm$ 0.0145 & 0.0485 $\pm$ 0.012 \\
ReCon \cite{lu2024recon} & 0.2791 $\pm$ 0.012 & 0.4986 $\pm$ 0.031 & 0.2811 $\pm$ 0.012 & 0.2215 $\pm$ 0.013 & 0.6681 $\pm$ 0.010 & 0.0376 $\pm$ 0.020 \\
SeedSelect \cite{samuel2024generating} & 0.1976 $\pm$ 0.010 & 0.4801 $\pm$ 0.021 & 0.2789 $\pm$ 0.011 & 0.2013 $\pm$ 0.032 & 0.6705 $\pm$ 0.012 & 0.0365 $\pm$ 0.020 \\
\hline

\multicolumn{7}{c}{\textbf{LLM-Enhanced Methods}} \\
\hline
LMD \cite{lian2023llm} & 0.2011 $\pm$ 0.012 & 0.4328 $\pm$ 0.010 & 0.2774 $\pm$ 0.045 & 0.2108 $\pm$ 0.031 & 0.6356 $\pm$ 0.015 & 0.0368 $\pm$ 0.013 \\
LLM-Blueprint \cite{gani2023llm} & 0.1901 $\pm$ 0.010 & 0.4561 $\pm$ 0.013 & 0.2705 $\pm$ 0.012 & 0.1943 $\pm$ 0.011 & 0.6513 $\pm$ 0.011 & 0.0312 $\pm$ 0.010 \\
Ranni \cite{Feng_2024_CVPR} & 0.1897 $\pm$ 0.011 & 0.4289 $\pm$ 0.011 & 0.2619 $\pm$ 0.022 & 0.2002 $\pm$ 0.012 & 0.6568 $\pm$ 0.010 & 0.0313 $\pm$ 0.021 \\
\hline

\multicolumn{7}{c}{\textbf{Ours (RAVEL)}} \\
\hline
\textbf{Flux + Ours} & \textbf{0.3311 $\pm$ 0.017} & \textbf{0.6810 $\pm$ 0.033} & \textbf{0.5824 $\pm$ 0.028} & \textbf{0.4220 $\pm$ 0.024} & \textbf{0.2881 $\pm$ 0.019} & \textbf{0.0867 $\pm$ 0.010} \\
\textbf{SDXL + Ours} & \textbf{0.3011 $\pm$ 0.016} & \textbf{0.6856 $\pm$ 0.032} & \textbf{0.5691 $\pm$ 0.027} & \textbf{0.4115 $\pm$ 0.022} & \textbf{0.3225 $\pm$ 0.020} & \textbf{0.0752 $\pm$ 0.009} \\
\textbf{DALL-E + Ours} & \textbf{0.3356 $\pm$ 0.018} & \textbf{0.7456 $\pm$ 0.035} & \textbf{0.6421 $\pm$ 0.029} & \textbf{0.4478 $\pm$ 0.026} & \textbf{0.2661 $\pm$ 0.017} & \textbf{0.0932 $\pm$ 0.010} \\
\hline
\end{tabular}%
}
\label{tab:image-generation-clip}
\end{table*}

\subsection{Quantitative Results} 
\noindent We comprehensively evaluate \texttt{RAVEL} across three tasks: rare concept generation, knowledge-grounded editing, and iterative self-correction. To assess low-level alignment and perceptual quality, we report standard image-text metrics across all tasks and backbones (Flux, SDXL, DALL-E 3). See tables. \ref{tab:image-generation-clip} and \ref{tab:self-correction-clip}. For high-level semantic alignment, we benchmark on \textit{T2I-CompBench} and three new datasets introduced in this work - \textit{MythoBench}, \textit{Rare-Concept-1K}, and \textit{NovelBench}, each targeting rare or culturally grounded concepts with varying difficulty (see table. \ref{tab:llmjudge-benchmarks}. We evaluate these using our LLM-as-a-Judge framework with interpretability-aware metrics: \textit{Attribute Accuracy Index (AAI)}, \textit{Context Relevance Score (CRS)}, \textit{Intent Representation Metric (IRM)}, and \textit{Visual Fidelity Score (VFS)}, implemented via the GEval library \cite{gralinski2019geval}.

\subsection{T2I-CompBench Results}

We compare \texttt{RAVEL} against state-of-the-art methods in RAG-based generation \cite{lu2024recon, shalev2025imagerag, blattmann2022ragdiffusion, samuel2024generating}, LLM-enhanced prompting \cite{gani2023llm, lian2023llm, Feng_2024_CVPR}, model-based editing \cite{wang2024genartist, Zhang_2024_CVPR}, and self-correction \cite{wu2024self, zhuo2025reflection}. Across all benchmarks, \texttt{RAVEL} demonstrates consistent gains in alignment, controllability, and fidelity, particularly for abstract, rare, and culturally nuanced concepts. Using G-Eval \cite{liu2023g}, we further evaluate our KG-RAG pipeline’s retrieval and generation quality through context relevance and answer accuracy metrics. 

Table   \ref{tab:image-generation-clip} shows the performance comparison of our approach across various text-image alignment, semantic coherence, and perceptual quality metrics against Flux, SDXL, and DALL-E 3 over a set of 500 images. Traditional models such as CLIP often struggle with fine-grained, domain-specific attributes of rare concepts, making direct prompt-to-image alignment metrics less informative. To address this, we compute text-image alignment metrics such as CLIP-T and SigLIP by comparing the text embeddings of a character’s ground-truth contextual description with the embeddings of the generated image. This provides deeper insight into how well the generated images reflect the expected characteristics of rare entities. Similarly, we assess image-image alignment using CLIP-I, DINO v2, and LPIPS by comparing generated images against reference images from authoritative sources.

\noindent To further enrich our evaluation, we leverage the BLIP-VQA model to generate structured descriptions of key aspects of the generated images - such as appearance traits, background, and accessories. We then measure the semantic similarity between these descriptions and the ground-truth context, yielding BLIP scores that offer an additional layer of alignment assessment. This approach is applied to both baseline model outputs and images generated using our method. Our findings indicate a substantial improvement in both text-image and image-image alignment metrics, demonstrating that our approach effectively grounds image generation in precise, character-specific contexts, mitigating the ambiguity inherent in traditional T2I models. Furthermore, \texttt{RAVEL} outperforms other RAG-based and LLM-enhanced methods as shown in the Table. \ref{tab:image-generation-clip}. This could be attributed to the fact that unlike these methods, our approach works even in the absence of visual priors. Also, KG-based RAG provides significantly superior and complete attribute retrieval, enhancing the quality of prompts for the T2I generation. 
Structured attribute retrieval ensures tight attribute clustering \& coverage, demonstrating the benefits of our graph-driven relational guidance, even in data sparse and complex concept structure scenarios.

\noindent This trend of improvement in metrics extends to self-correction, as shown in Table \ref{tab:self-correction-clip}, where our SRD algorithm iteratively improves contextual alignment with each cycle of refinement, taking less time compared to SOTA methods. Repetitive generation and averaging of metrics using random resampling, and other self-correction methods still depend on the models' pretrained knowledge, hence failing for cases when the model is not trained on those images or visual priors are absent. Furthermore, our approach significantly improves the ability of ControlNet to perform contextually accurate and narratively rich edits by incorporating relational details of rare characters as seen in Table \ref{tab:editing-clip}. While other metrics show similar trends, note that the LPIPS metric represents an inherent trade-off when it comes to editing. While we want to lower LPIPS to enhance perceptual similarity of the edited images with the ground truth images, we also want to ensure meaningful modifications to the image due to editing, hence increasing LPIPS. Our approach carefully balances this trade-off. While \texttt{RAVEL}'s CLIP-I, CLIP-T remain higher and LPIPS lower for image generation across CLIP-LPIPS plots, our method successfully takes up the upper-left corner for editing, achieving high CLIP similarity scores and balanced LPIPS scores. 

\begin{table*}[t]
\caption{\textbf{Self-Correction Performance Comparison.} RAVEL with SRD outperforms prior baselines in accuracy, alignment, and efficiency.}
\centering
\footnotesize
\label{self-correction-clip}
\renewcommand{\arraystretch}{1.2}

\resizebox{\textwidth}{!}{%
\begin{tabular}{l|c|c|c|c|c|c|c|c}
\hline
\textbf{Method} & \textbf{Rounds} & \textbf{CLIP-T} $\uparrow$ & \textbf{CLIP-I} $\uparrow$ & \textbf{DINO} $\uparrow$ & \textbf{BLIP-VQA} $\uparrow$ & \textbf{LPIPS} $\downarrow$ & \textbf{SigLIP} $\uparrow$ & \textbf{Time (s)} $\downarrow$ \\
\hline
\multicolumn{9}{c}{\textbf{Base Models (no correction)}} \\
\hline
Flux & 0 & 0.1871 $\pm$ 0.012 & 0.418 $\pm$ 0.0018 & 0.2871 $\pm$ 0.019 & 0.2278 $\pm$ 0.018 & 0.5802 $\pm$ 0.031 & 0.0634 $\pm$ 0.007 & 62.5 \\
SDXL & 0 & 0.1842 $\pm$ 0.013 & 0.297 $\pm$ 0.0013 & 0.2619 $\pm$ 0.017 & 0.1994 $\pm$ 0.016 & 0.6234 $\pm$ 0.034 & 0.0335 $\pm$ 0.005 & 8.3 \\
DALL-E 3 & 0 & 0.2201 $\pm$ 0.014 & 0.4802 $\pm$ 0.027 & 0.3667 $\pm$ 0.022 & 0.2789 $\pm$ 0.020 & 0.5677 $\pm$ 0.030 & 0.0689 $\pm$ 0.008 & 34.2 \\
\hline
\multicolumn{9}{c}{\textbf{Test-Time Scaling (random resampling)}} \\
\hline
Flux + Resample (N=4) & – & 0.1908 $\pm$ 0.003 & 0.4121 $\pm$ 0.014 & 0.2278 $\pm$ 0.013 & 0.2281 $\pm$ 0.012 & 0.5640 $\pm$ 0.012 & 0.0568 $\pm$ 0.003 & 245.8 \\
Flux + Resample (N=8) & – & 0.1893 $\pm$ 0.002 & 0.4050 $\pm$ 0.012 & 0.2787 $\pm$ 0.010 & 0.2127 $\pm$ 0.001 & 0.5521 $\pm$ 0.014 & 0.0562 $\pm$ 0.004 & 493.6 \\
\hline
\multicolumn{9}{c}{\textbf{Other Self-Correction Methods}} \\
\hline
SLD \cite{wu2024self} & 2–3 & 0.2510 $\pm$ 0.010 & 0.4822 $\pm$ 0.022 & 0.3561 $\pm$ 0.019 & 0.2891 $\pm$ 0.021 & 0.5711 $\pm$ 0.026 & 0.0623 $\pm$ 0.004 & 178.7 \\
ReflectionFlow \cite{zhuo2025reflection} & 16 & 0.2472 $\pm$ 0.005 & 0.5219 $\pm$ 0.013 & 0.3891 $\pm$ 0.011 & 0.3127 $\pm$ 0.015 & 0.5234 $\pm$ 0.010 & 0.0781 $\pm$ 0.005 & 312.3 \\
\hline
\multicolumn{9}{c}{\textbf{RAVEL with SRD (Ours)}} \\
\hline
Flux + SRD R1 & 1 & 0.3381 $\pm$ 0.014 & 0.7110 $\pm$ 0.032 & 0.6124 $\pm$ 0.008 & 0.4320 $\pm$ 0.014 & 0.2881 $\pm$ 0.019 & 0.0962 $\pm$ 0.021 & 155.5 \\
Flux + SRD R2 & 2 & \textbf{0.3379 $\pm$ 0.015} & \textbf{0.7515 $\pm$ 0.030} & \textbf{0.6412 $\pm$ 0.026} & \textbf{0.4413 $\pm$ 0.023} & \textbf{0.2945 $\pm$ 0.018} & \textbf{0.1302 $\pm$ 0.012} & 267.8 \\
SDXL + SRD R1 & 1 & 0.3121 $\pm$ 0.011 & 0.7156 $\pm$ 0.012 & 0.5811 $\pm$ 0.021 & 0.4215 $\pm$ 0.022 & 0.3255 $\pm$ 0.010 & 0.0912 $\pm$ 0.009 & 46.2 \\
SDXL + SRD R2 & 2 & \textbf{0.3205 $\pm$ 0.014} & \textbf{0.7334 $\pm$ 0.031} & \textbf{0.5978 $\pm$ 0.025} & \textbf{0.4278 $\pm$ 0.021} & \textbf{0.3301 $\pm$ 0.011} & \textbf{0.1087 $\pm$ 0.011} & 88.3 \\
DALL-E 3 + SRD R1 & 1 & 0.3451 $\pm$ 0.011 & 0.7512 $\pm$ 0.015 & 0.6521 $\pm$ 0.020 & 0.4518 $\pm$ 0.022 & 0.2682 $\pm$ 0.010 & 0.1021 $\pm$ 0.010 & 75.9 \\
DALL-E 3 + SRD R2 & 2 & \textbf{0.3678 $\pm$ 0.016} & \textbf{0.7801 $\pm$ 0.033} & \textbf{0.6933 $\pm$ 0.028} & \textbf{0.4821 $\pm$ 0.025} & \textbf{0.2787 $\pm$ 0.016} & \textbf{0.1388 $\pm$ 0.013} & 158.1 \\
\hline
\end{tabular}%
}
\label{tab:self-correction-clip}
\end{table*}

Since conventional metrics may struggle to capture nuanced and context-sensitive details essential to distinguish complex attributes in rare characters, we further integrate the `LLM as a Judge' \cite{zheng2023llm_as_a_judge} framework. This framework, which has demonstrated near-human reasoning capabilities, enables dynamic evaluation of generated images based on the narrative context retrieved from our knowledge graph. We sample 100 rare image prompts for each of the four benchmarks and evaluate them using the four metrics that we curate: \texttt{Attribute Accuracy}, \texttt{Context Relevance}, \texttt{Visual Fidelity}, and \texttt{Intent Representation}. The LLM Judge is tasked with assigning a score between 0 and 1 for each metric, ranging from poor to good, to evaluate the quality of the image based on these criteria. Overall Alignment is calculated using the weighted scores of the metrics with [0.6, 0.15, 0.1, 0.15] weights for the corresponding metrics to prioritize accurate character depictions without compromising visual quality and narrative coherence.
Table \ref{tab:llmjudge-benchmarks} shows that our method significantly enhances context-aware character depiction across all baseline models and benchmarks, achieving an average improvement of over \textbf{30\%} in attribute accuracy and overall alignment.

\begin{table*}[t]
\caption{Image editing performance comparison of ControlNet with and without our method. *We want to lower LPIPS as we are performing disentangled edits but not too low because we want to meaningfully modify the image so a moderate value indicates good performance.}
\centering
\footnotesize
\setlength{\tabcolsep}{6pt}
\renewcommand{\arraystretch}{1.2}
\begin{tabular}{l|c|c|c|c|c|c}
\toprule
\textbf{Method} & \textbf{CLIP-T} $\uparrow$ & \textbf{CLIP-I} $\uparrow$ & \textbf{DINO} $\uparrow$ & \textbf{BLIP-VQA} $\uparrow$ & \textbf{LPIPS} $\downarrow$* & \textbf{SigLIP} $\uparrow$ \\
\midrule
\multicolumn{7}{c}{\textbf{Instruction-Based Editing}} \\
\hline
InstructPix2Pix \cite{brooks2023instructpix2pix}        & 0.2912 $\pm$ 0.012 & 0.9014 $\pm$ 0.018 & 0.8812 $\pm$ 0.014 & 0.3685 $\pm$ 0.011 & 0.0894 $\pm$ 0.006 & 0.1011 $\pm$ 0.008 \\
Prompt2Prompt \cite{hertz2022prompt}         & 0.2756 $\pm$ 0.010 & 0.9120 $\pm$ 0.020 & 0.8898 $\pm$ 0.015 & 0.3723 $\pm$ 0.012 & 0.0825 $\pm$ 0.005 & 0.0964 $\pm$ 0.007 \\
DiffEdit \cite{couairon2022diffedit}              & 0.3033 $\pm$ 0.011 & 0.9187 $\pm$ 0.016 & 0.9021 $\pm$ 0.013 & 0.3792 $\pm$ 0.010 & 0.0762 $\pm$ 0.004 & 0.1073 $\pm$ 0.006 \\
\midrule
\multicolumn{7}{c}{\textbf{Model-Based}} \\
\hline
ControlNet (base)      & 0.3179 $\pm$ 0.011 & 0.9810 $\pm$ 0.020 & 0.9392 $\pm$ 0.001 & 0.3682 $\pm$ 0.021 & 0.0638 $\pm$ 0.010 & 0.1102 $\pm$ 0.001 \\
HIVE \cite{Zhang_2024_CVPR}                  & 0.3015 $\pm$ 0.013 & 0.9265 $\pm$ 0.017 & 0.9110 $\pm$ 0.012 & 0.3756 $\pm$ 0.011 & 0.2313 $\pm$ 0.004 & 0.1155 $\pm$ 0.007 \\
GenArtist \cite{wang2024genartist}              & 0.3078 $\pm$ 0.014 & 0.9342 $\pm$ 0.018 & 0.9159 $\pm$ 0.013 & 0.3768 $\pm$ 0.012 & 0.2274 $\pm$ 0.005 & 0.1183 $\pm$ 0.006 \\
\midrule
\textbf{ControlNet + RAVEL (Ours)} & \textbf{0.3578} $\pm$ 0.021 & \textbf{0.9401} $\pm$ 0.006 & \textbf{0.9221} $\pm$ 0.022 & \textbf{0.4291} $\pm$ 0.001 & \textbf{0.2835} $\pm$ 0.002 & \textbf{0.1279} $\pm$ 0.012 \\
\bottomrule
\end{tabular}
\label{tab:editing-clip}
\end{table*}

\noindent \textbf{User Study:} We conducted a controlled prolific.com user study (N=60) evaluating Attribute Accuracy, Visual Fidelity, Contextual Relevance, and Intent Representation across 100 generated images. Across 25 judgment tasks per participant, 91.93\% of choices preferred RAVEL-enhanced outputs.

\begin{table}[t]
\vspace{-2mm}
\centering
\scriptsize
\caption{\textbf{User Study Results.} Percentage of user preferences for each T2I model with and without \texttt{RAVEL}.}
\label{tab:user-study}
\renewcommand{\arraystretch}{1.05}
\setlength{\tabcolsep}{4.5pt}

\resizebox{\columnwidth}{!}{%
\begin{tabular}{lcccc}
\toprule
\textbf{Model} & \textbf{Attr. Acc.} & \textbf{Vis. Fidelity} & \textbf{Context Rel.} & \textbf{Intent Repr.} \\
\midrule
Flux (baseline)           & 58.2\% & 60.1\% & 56.3\% & 55.0\% \\
Flux + \texttt{RAVEL}     & \textbf{92.4\%} & \textbf{90.3\%} & \textbf{91.2\%} & \textbf{89.7\%} \\
\midrule
SDXL (baseline)           & 61.5\% & 63.8\% & 59.2\% & 57.9\% \\
SDXL + \texttt{RAVEL}     & \textbf{90.1\%} & \textbf{88.6\%} & \textbf{90.4\%} & \textbf{87.8\%} \\
\midrule
DALL·E 3 (baseline)       & 60.3\% & 62.0\% & 58.1\% & 56.5\% \\
DALL·E 3 + \texttt{RAVEL} & \textbf{91.6\%} & \textbf{89.4\%} & \textbf{90.8\%} & \textbf{88.9\%} \\
\bottomrule
\end{tabular}
}
\vspace{-2mm}
\end{table}

\section{Ablation Studies}
\label{sec:ablation}

To evaluate each component in our framework, we conduct comprehensive ablation studies on a sample of 100 images across image generation, self-correction, and editing. We evaluate how the four contextually aware alignment metrics - Attribute Accuracy, Context Relevance, Visual Fidelity, and Intent Representation defined using the LLMJudge framework vary with retrieval, prompting strategies. 

\begin{figure}[t!]
    \centering
    \includegraphics[width=\columnwidth]{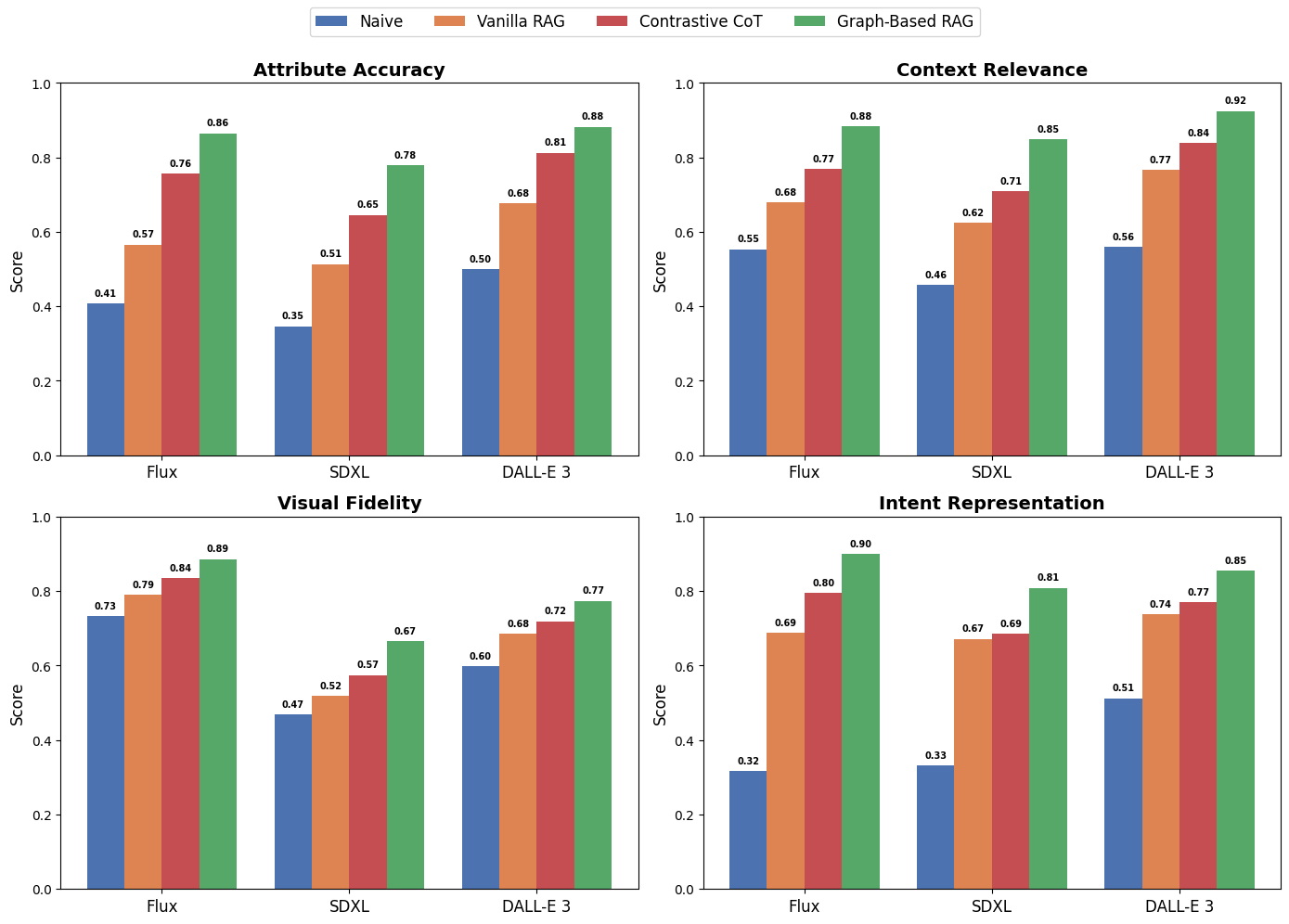} %
    \caption{Ablation study demonstrating how different retrieval and prompting strategies contribute to \texttt{RAVEL}'s effectiveness in enhancing T2I models.}
    \label{fig:image-generation-ablation}
\end{figure}

\noindent \textbf{Image Generation:} We compare vanilla RAG, Contrastive CoT prompting, and our graph-based retrieval across Flux, SDXL, and DALL·E (Figure~\ref{fig:image-generation-ablation}). All methods improve generation quality, with vanilla RAG boosting Flux’s attribute accuracy by 15\% and intent representation by 35\%. Contrastive CoT further enhances all four metrics by structuring multi-faceted context. Our graph-based retrieval outperforms both by providing rich, interconnected context that aligns images with both explicit and implicit attributes, yielding the highest gains across all models and metrics.

\noindent \textbf{Self-Correction and Editing:} We evaluate our SRD self-correction algorithm over two refinement iterations across four metrics and multiple T2I models. Each iteration yields clear gains by progressively strengthening weak or missing attributes through structured, context-aware prompt updates. Naïve baselines perform poorly due to their lack of contextual grounding, while our graph-based RAG initialization already provides strong improvements by supplying coherent, attribute-complete context. SRD further reduces hallucinations, sharpens attribute fidelity, and improves semantic alignment across models. For editing, standard ControlNet often introduces generic or misaligned modifications when applied to rare concepts. Integrating our method yields context-aware edits that preserve core attributes, substantially improving performance on all four evaluation metrics.

\section{Conclusion}
\label{sec:conclusion}
We present \texttt{RAVEL}, a graph-based RAG framework that enhances T2I diffusion models by addressing limitations in representing rare, domain-specific concepts. By leveraging structured knowledge graphs, \texttt{RAVEL} improves accuracy, consistency, and cultural alignment without requiring fine-tuning. It supports context-aware editing and iterative self-correction, enabling faithful generations and advanced relational modifications. Compatible with models like SDXL, Flux, and ControlNet, \texttt{RAVEL} offers a versatile solution for controllable, knowledge-grounded image generation.

\clearpage

{
    \small
    \bibliographystyle{ieeenat_fullname}
    \bibliography{main}
}



\end{document}